\documentclass[letterpaper]{article} % DO NOT CHANGE THIS
\usepackage{aaai24}  % DO NOT CHANGE THIS
\usepackage{times}  % DO NOT CHANGE THIS
\usepackage{helvet}  % DO NOT CHANGE THIS
\usepackage{courier}  % DO NOT CHANGE THIS
\usepackage[hyphens]{url}  % DO NOT CHANGE THIS
\usepackage{graphicx} % DO NOT CHANGE THIS
\usepackage{natbib}  % DO NOT CHANGE THIS AND DO NOT ADD ANY OPTIONS TO IT
\usepackage{caption} % DO NOT CHANGE THIS AND DO NOT ADD ANY OPTIONS TO IT
\usepackage{algorithm}
\usepackage{algorithmic}

\usepackage{amsmath}
\usepackage{amsfonts}
\usepackage{booktabs}
\usepackage{tabularx}
\usepackage{float}

\usepackage{newfloat}
\usepackage{listings}
\DeclareCaptionStyle{ruled}{labelfont=normalfont,labelsep=colon,strut=off} % DO NOT CHANGE THIS
\floatstyle{ruled}
\newfloat{listing}{tb}{lst}{}
\floatname{listing}{Listing}
\nocopyright
\title{Causal Intersectionality and Dual Form of Gradient Descent for Multimodal Analysis: a Case Study on Hateful Memes}
\author{
    Yosuke Miyanishi\textsuperscript{\rm 1}, Minh Le Nguyen\textsuperscript{\rm 1}\\
}
\affiliations{
    \textsuperscript{\rm 1}Japan Advanced Institute of Science and Technology\\
    1-1 Asahi-dai, Nomi-shi, Ishikawa, Japan\\
    \{yosuke.miyanishi, nguyenml\}@jaist.ac.jp\\
}

\begin{document}

\maketitle

\begin{abstract}
Amidst the rapid expansion of Machine Learning (ML) and Large Language Models (LLMs), understanding the semantics within their mechanisms is vital. Causal analyses define semantics, while gradient-based methods are essential to eXplainable AI (XAI), interpreting the model's 'black box'. Integrating these, we investigate how a model's mechanisms reveal its causal effect on evidence-based decision-making. Research indicates intersectionality - the combined impact of an individual's demographics - can be framed as an Average Treatment Effect (ATE). This paper demonstrates that hateful meme detection can be viewed as an ATE estimation using intersectionality principles, and summarized gradient-based attention scores highlight distinct behaviors of three Transformer models. We further reveal that LLM Llama-2 can discern the intersectional aspects of the detection through in-context learning and that the learning process could be explained via meta-gradient, a secondary form of gradient. In conclusion, this work furthers the dialogue on Causality and XAI. Our code is available online (see \textit{External Resources} section).
\end{abstract}
\section{Introduction}
The domain of causality offers profound insights into the data generation processes, revealing the intricate architecture of the problems at hand. A meticulous examination of these generative processes is indispensable for deep comprehension of phenomena with significant social implications. This paper is dedicated to conducting a rigorous case study in this vein, bridging the gap between the foundational principles of science and the cutting-edge capabilities of Machine Learning (ML) technologies.\\
EXplainable Artificial Intelligence (XAI) emerges as a critical paradigm in shedding light on ML models' often opaque inner workings. While previous research has ventured into various domains, the application of XAI principles to causal analysis remains scarcely explored. By integrating causality and XAI, this study aims to enrich our understanding of social phenomena and how they are reflected in state-of-the-art (SOTA) ML models facing the representation of the phenomena.\\
The rise of online hate speech, especially hateful memes (Fig. 1, top) —comprising both text and image, has prompted significant research. While multimodal ML algorithms have seen substantial improvements, efforts focus more on benchmarking and maximizing performance, including the Hateful Memes Challenge competition \cite{kielaHatefulMemesChallenge2020}, rather than applying XAI methods. Existing approaches also lack a focus on causal architecture. This study defines hateful meme detection as an Average Treatment Effect (ATE) estimation problem for input data modalities (image and text) and examines the effects through the prism of XAI.

\begin{figure}[!ht]
    \begin{center}
    %\fbox{\parbox{6cm}{
    %This is a figure with a caption.}}
    % old picture \includegraphics[scale=0.5]{lrec2020W-image1.eps}
    \includegraphics[scale=0.5]{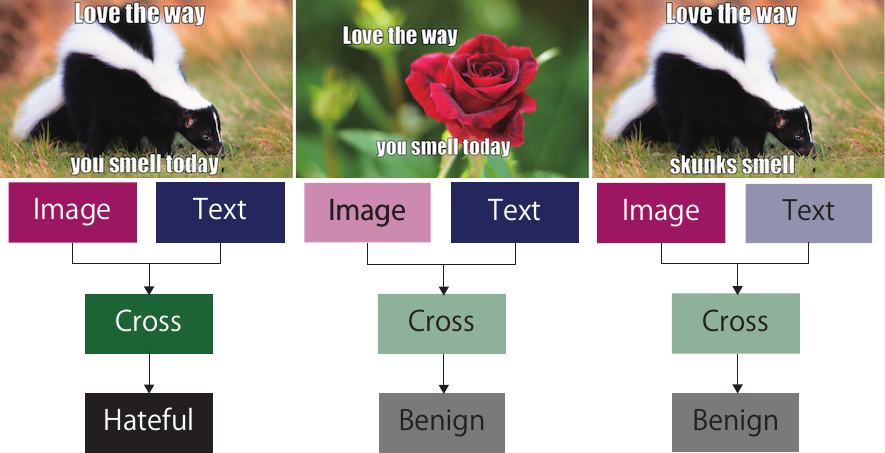} 

    \caption{Visualization of a hateful meme and its corresponding confounders. (top) Meme samples and (bottom) their directed acyclic graph representation. (left) A hateful meme highlights cross-modal interactions between its image and text components that contribute to its hatefulness. (middle) The image benign confounder showcases original text and a non-hateful image, resulting in reduced cross-modal interactions and a benign classification. (right) The text benign confounder comprises an original image and non-hateful text. Note: The samples depicted are illustrative and do not exist in the dataset. ©Getty Images}
    \label{fig.1}
    \end{center}
\end{figure}

Intersectionality\footnote{Oxford Dictionary}, or \textit{the network of connections between social categories such as race, class, and gender, especially when this may result in additional disadvantage or discrimination}, acts as a bridge between ML and social science. Though broadly applied in social science and used for debiasing in ML, its wider applications are limited. How can we use this concept as a \textit{generalized} tool for broader problems? Motivated by this question, this paper proposes reframed intersectionality, explores whether causally formalized intersectionality can address a broader range of problems, and evaluates \textit{inductive} bias in ML models.\\ 
Furthermore, the excellence of Large Language Models (LLMs) across various benchmarks has been showcased, particularly in few-shot learning. The concept of in-context learning presents a promising avenue, but its formal causal evaluation is limited. Here, we address this problem.\\
Our contribution could be summarized as:
\begin{itemize}
\item Formalization of hateful meme detection as an intersectional causal effect estimation problem, allowing performance assessment based on data generation process.
\item Introduction of reframed causal intersectionality to evaluate inductive bias, marking a step towards broader applications, including demographics-nationality intersectionality, financial inclusion, and clinical diagnosis.
\item Demonstration that attention attribution scores \cite{haoSelfAttentionAttributionInterpreting2021} divided by modality interaction describe the causal effect accurately, unlike non-divided scores. This finding opens doors for causal explainability in multimodal settings \cite{liuCausalReasoningMeets2022}.
\item Pioneering formal and empirical analysis of LLM's meta-optimization process in the multimodal in-context setting.
\end{itemize}

\section{Related Work}
\subsection{Causal ML and XAI}
Causal Inference (CI) occupies a pivotal role in the elucidation of social phenomena and the interpretation of intervention outcomes. It bifurcates into two primary methodologies: the graphical and structural schemas for modeling reality \cite{pearlCAUSALITYMODELSREASONING2001}, alongside the framework for potential outcome prediction \cite{rubinObjectiveCausalInference2008}. CI's utility spans a diverse array of sectors, including but not limited to, medicine \cite{vlontzosReviewCausalityLearning2022}, manufacturing \cite{vukovicCausalDiscoveryManufacturing2022}, and the social sciences \cite{kinshuksenguptaCausalEffectRacial2021}, guiding the interpretation of data within these fields. Furthermore, CI principles have been applied within Machine Learning (ML) and its allied disciplines, giving rise to the subfield of Causal ML. Causal ML encompasses research into natural language processing \cite{yangSurveyExtractionCausal2022}, hate speech detection \cite{chakrabortyNippingBudDetection2022}, and the study of image-text multimodality \cite{sanchezCausalMachineLearning2022}. Notably, the theoretical underpinnings of hateful memes, as a convergence of these interests, remain underexplored. Our research attempts to map out graphical and formal representations of the causal structures underlying hateful memes.\\
In conjunction, EXplainable Artificial Intelligence (XAI) \cite{speithReviewTaxonomiesExplainable2022,joshiReviewExplainabilityMultimodal2021,barredoarrietaExplainableArtificialIntelligence2020} has sought to demystify the internal mechanisms of ML models. XAI's domain of inquiry extends across various fields, including medicine \cite{holzingerExplainableAIMultiModal2021} and energy \cite{machlevExplainableArtificialIntelligence2022}, with a particular focus on both model-agnostic \cite{sundararajanAxiomaticAttributionDeep2017,gaurSemanticsBlackBoxCan2021,marcosSemanticallyInterpretableActivation2019} and model-specific \cite{haoSelfAttentionAttributionInterpreting2021,holzingerMultimodalCausabilityGraph2021a} evaluations. However, the intersection of causality with XAI remains nascent. This study investigates XAI's capability in assessing attributions to causality metrics, emphasizing gradient-based interpretations as central to XAI endeavors.\\
Since ML models typically minimize the gradient for optimization, components with steep gradients toward the model’s decision-making are considered crucial. The gradient-based XAI approach \cite{selvarajuGradCAMVisualExplanations2017}, often model-specific, finds pertinent application in the analysis of Transformers \cite{vaswaniAttentionAllYou2017}, which underpin most SOTA models in natural language processing (NLP). Here, quantifying the attribution of attention matrix weights via the gradient emerges as a direct method \cite{haoSelfAttentionAttributionInterpreting2021}. Our research proposes both theoretical and empirical advancements in causal analysis, leveraging this gradient-based methodology to enhance understanding and interpretation within the causality domain.

\subsection{Intersectionality}
Intersectionality, a bias indicator of multiple demographics within various domains, has inspired a few causal analyses \cite{yangCausalIntersectionalityFair2021,brightCausallyInterpretingIntersectionality2016}. While XAI techniques have been used to alleviate its negative impact in ML literature \cite{lalorBenchmarkingIntersectionalBiases2022,simateleFinancialInclusionIntersectionality2022}, our work redefines intersectionality for broader problems, and pioneers the quantification of inductive intersectional bias.

\subsection{Hate Speech and Hateful Memes}
Hate speech and hateful memes have attracted substantial ML research attention, involving various models \cite{dasDetectingHateSpeech2020,lippeMultimodalFrameworkDetection2020a} and datasets \cite{kielaHatefulMemesChallenge2020,degibertHateSpeechDataset2018,davidsonAutomatedHateSpeech2017,sabatHateSpeechPixels2019,suryawanshiMultimodalMemeDataset2020}. Previous analytical works have focused on racial bias \cite{kinshuksenguptaCausalEffectRacial2021,sharmaDISARMDetectingVictims2022}, virality \cite{lingDissectingMemeMagic2021,chakrabortyNippingBudDetection2022}, and propaganda techniques \cite{dimitrovDetectingPropagandaTechniques2021}, and a few have applied XAI methods \cite{caoDisentanglingHateOnline2021,heeExplainingMultimodalHateful2022,deshpandeInterpretableApproachHateful2021}. This study builds upon these by formalizing hateful meme detection as a causal effect estimation problem and emphasizing the importance of modality interaction.

\subsection{LLM}
Large Language Models (LLMs), known for their powerful in-context few-shot learning capabilities \cite{brownLanguageModelsAre2020} in various NLP and multimodal tasks, are emerging as significant tools \cite{zhaoSurveyLargeLanguage2023}. To understand their inner workings, meta-gradient, or the update of the attention weights as a secondary form of the gradient, explains in-context learning empirically \cite{coda-fornoMetaincontextLearningLarge2023,chenMetalearningLanguageModel2022} and theoretically \cite{daiWhyCanGPT2023a,vonoswaldTransformersLearnInContext2023}.\\
However, understanding LLM's causal power remains a complex and emerging area of study \cite{banQueryToolsCausal2023,kicimanCausalReasoningLarge2023}.  Moreover, unlike the traditional classifier-attached-encoder model (e.g. BERT \cite{devlinBERTPretrainingDeep2019}) with \textit{predicted probability}, how to estimate the causal effect of chatbot-style LLM and analyze its inner workings quantitatively remains elusive.
This study demonstrates that a dedicated task design could be used to estimate LLM's causal effect and that a meta-gradient could explain its inner workings concerning that effect.

\section{Methodology}
\subsection{Background}
\subsubsection{Average Treatment Effect}
The \textit{Average Treatment Effect (ATE)} \cite{rubinObjectiveCausalInference2008} is a key metric for assessing causal impacts. It represents the average difference in outcomes between treated and untreated groups. This measurement facilitates a standardized evaluation of causal effects across diverse scenarios. For a binary treatment $BT \in (0,1)$ yielding outcome $\theta_{BT}$, $ATE$ is defined as: 
\begin{equation} \label{eq:1}
    ATE = \theta_1 - \theta_0
\end{equation}
This research repurposes intersectional $ATE$ to evaluate hateful meme detection models.

\subsubsection{Causal Intersectionality}
Building upon the textual definition, causal intersectionality \cite{brightCausallyInterpretingIntersectionality2016} challenges the simplistic aggregation of individual demographic effects. Instead, it highlights the complex, synergistic interactions between multiple demographic factors, acknowledging the nuanced dynamics that influence causal relationships in social studies. Defining causal intersectionality (Eq. 2) involves binary vectors for two demographics (e.g., gender $D_1$ and color $D_2$) marked as $D = \{D_1, D_2\}$, and the outcome $\theta$. 
\begin{equation} \label{eq:2}
\theta_D \neq \sum_i \theta_{D_i}
\end{equation}
Using this causality structure, we assess multimodal models within a causal context.

\subsubsection{Attention Attribution Score}
Simply put, \textit{attention attribution score} quantifies the contribution of attention weights to the model's decision-making. Initially, a seminal work \cite{sundararajanAxiomaticAttributionDeep2017} introduced the \textit{integrated gradient} method for quantifying model component's attribution. This method calculates the contribution of specific model components based on the gradient's integral over that component. Upon this work, another study \cite{haoSelfAttentionAttributionInterpreting2021} reported its applicability to Transformer's attention weights, deriving attention attribution score ($attr$ herein). This approach proves critical for interpreting the Transformer's behavior, especially in understanding how the attention matrix influences model outputs. Given hyperparameter $\alpha$, $attr$ computes the integrated gradient for attention matrix $A$ relative to a Transformer's output $\theta$.
\begin{equation} \label{eq:3}
attr = A*\int_{\alpha=0}^1\frac{\partial \theta(\alpha A)}{\partial A}d\alpha
\end{equation}
We employ modality-wise averaged $attr$ differences as causal effect indicators.

\subsubsection{Learning Objectives: Binary Classifier vs LLM}
For the classifier-attached-encoder model, hateful meme detection aligns with binary classification. In summary, given \textit{both hateful and benign pairs}, the classifier tries to maximize the classification performance. With a ground-truth label $y_{gt}$ and a loss function $f_{loss}$, the learning objective when training a model $\theta$ is:
\begin{equation} \label{eq:4}
\mathop{\mathrm{argmin}}_\theta -\{y_{gt}f_{loss}(\theta)+(1-y_{gt})f_{loss}(1-\theta)\}
\end{equation}
On the other hand, in in-context learning, hateful samples and their counterpart confounders are presented to LLM \textit{in parallel}. This differs from the objective above in that the information of hateful samples is not used to handle confounders, and vice versa. For example, facing the hateful sample in the zero-shot setting, the meta-objective it is trying to meta-optimize to is:
\begin{equation} \label{eq:5}
\mathop{\mathrm{argmin}}_\theta -y_{gt}f_{loss}(\theta)
\end{equation}
This study delves into the meta-objective for LLM in a few-shot context for the equivalent comparison.

\subsubsection{Meta-Optimization in Few Shot Setting}
Meta-optimization, in the realm of in-context learning, mirrors gradient descent \cite{irieDualFormNeural2022,daiWhyCanGPT2023a}. Given query $q$ as an input, a dual learning process of linear attention $\theta$ - fine-tuning and in-context learning with attention update $\Delta A$ - is contrasted with $A_{ZSL}$, the weight in a zero-shot setting.
\begin{equation}\label{eq:6}
    \theta = (A_{ZSL}+\Delta A)q
\end{equation}
$\Delta A$ functions as a gradient variant named meta-gradient. Our extension encapsulates attention attribution and its subsequent in-context learning results.

\subsection{Proposed Methodology Overview}
Expanding on the original causal intersectionality (Eq. 2), we define intersectionality for text $T$ and image $I$ modalities. With $X_1=(T_1, I_1)$ indicating hateful content and two types of benign samples $X_0 \in \{(T_1, I_0), (T_0, I_1)\}$, the multimodal intersectionality is:
\begin{equation} \label{eq:7}
\theta_{X_1} \neq \sum_{X_0} \theta_{X_0}
\end{equation}
The remainder of this section elaborates on causal intersectionality in meme detection, $attr$-based modality assessment, and LLM evaluation, visualized in Fig. 2.

\begin{figure}[!ht]
    \begin{center}
    %\fbox{\parbox{6cm}{
    %This is a figure with a caption.}}
    % old picture \includegraphics[scale=0.5]{lrec2020W-image1.eps} 
    \includegraphics[scale=0.5]{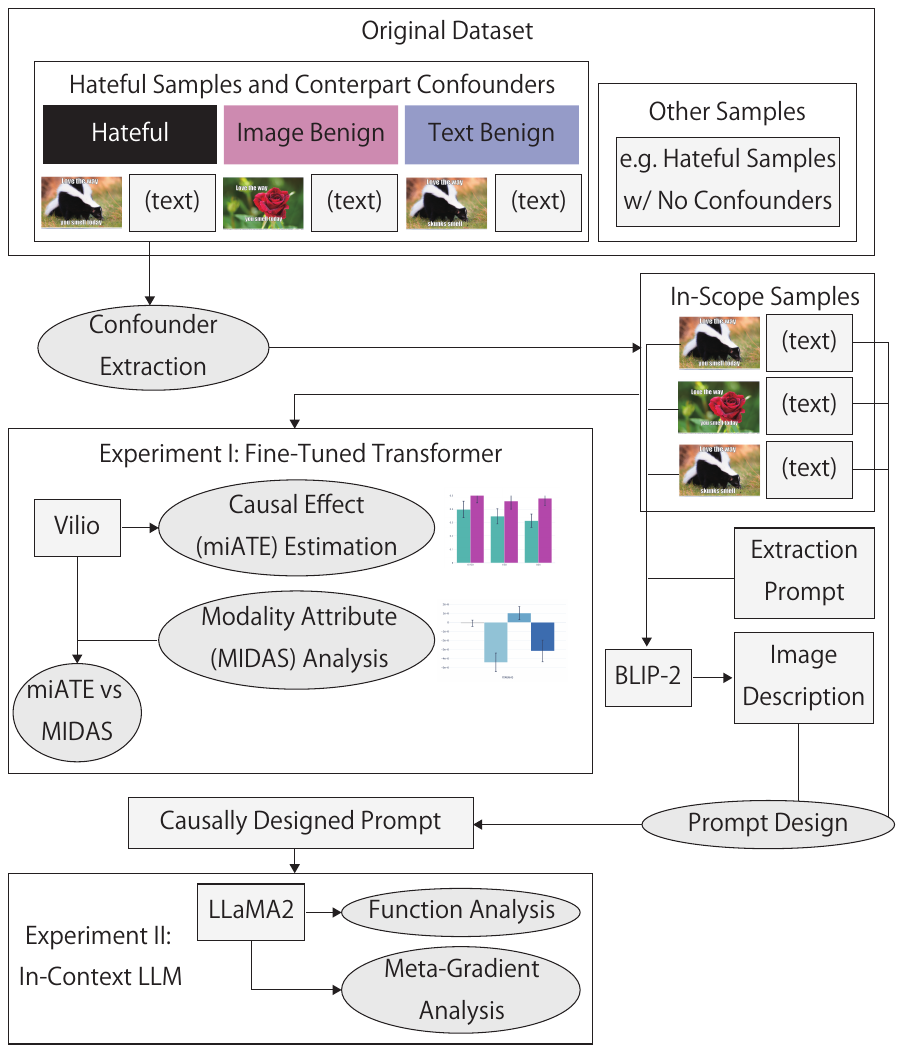} 

    \caption{A schematic overview of our proposed methodology. Rectangular boxes denote data or models, while circular shapes represent the processes involved.}
    \label{fig.2}
    \end{center}
\end{figure}

\subsection{Causal Multimodal Intersectionality}
\subsubsection{Intersectionality Reframed}
We broaden causal intersectionality, positing its utility beyond human demographics to include arbitrary components. This reframed intersectionality assesses interconnections between \textit{arbitrary categories such as user demographics or input modalities}, and how they amplify effects on significant issues - capturing indirect effects\footnote{Defined as an effect of two variables $X_1$ and $X_2$ to variable $X_3$ via another variable $Z$} in social contexts. This adaptation remains consistent with the original causal formalism (Eq. 2).

\subsubsection{Performance Measurement with Causal Intersectionality}
In the data generation process of hateful memes, a text and an image, which are benign in isolation, jointly produce hate. Applying demographic intersectionality concepts, we introduce the \textit{multimodal intersectional Average Treatment Effect (miATE)}.
\begin{equation} \label{eq:8}
miATE = \theta_{X_1} - \sum_{X_0} \theta_{X_0}
\end{equation}
Model performance is assessed considering differences in each modality. Hateful samples are categorized by original benign text confounders $T_0^{org}$ and image benign confounders $I_0^{org}$. The \textit{Confounder Extraction} section provides more details.

\subsection{Attention Attribution Score by Interaction}
We introduce the Modality Interaction Disentangled Attribution Score (MIDAS), which quantifies the contribution from various interaction types $t \in \{within\_image, within\_text, cross\_modal\}$ to a model's decision. Given an input $X_i=(T, I)$, with $IT$ denoting the number of elements for interaction type $t$, $MIDAS$ computes the modality-wise average $avg_t$ of the attention attribution score $attr^{X_i}$, analogous to $miATE$. Following the initial work on attention attribution score \cite{haoSelfAttentionAttributionInterpreting2021}, $MIDAS$ is calculated from the last hidden layer, excluding $[CLS]$ and $[SEP]$ tokens.
\begin{equation} \label{eq:9}
    \begin{aligned}
        MIDAS &= avg_t\{attr^{X_1} - \sum_{X_0} (attr^{X_0})\}\\
        avg_t(X) &= \frac{1}{IT} \sum^t X
    \end{aligned}
\end{equation}
Note that to mitigate the negative impact of class imbalance \cite{hossainMUTEMultimodalDataset2022}, $\theta$ and $attr$ is averaged per sample and confounder category ${(T_1,I_1),(T_1,I_0),(T_0,I_1)}$.

\subsection{Formal Relation between miATE and MIDAS}\label{relation}
We examine the relationship between $miATE$ and $MIDAS$ by proposing that $MIDAS$ can be perceived as an attention attribution to $miATE$. Given $G(A)$ as the one-step gradient for an attention matrix $A$, we see that $attr$ approximates the product of the gradient and the attention (Eq. 10 - true when $\alpha=1$). $MIDAS$ is expressed as the difference of that value between hateful and benign content (Eq. 11). Furthermore when $MIDAS$ is aggregated across $n$ samples that are representative of the entire dataset, it depicts the variation in attention expectancy normalized with the function $\mathcal{N}$ across these samples, as shown in Eq. 11.  In essence, we propose that $MIDAS$ acts as an attention-focused representation of the model's causal effect, i.e. $miATE$.
\begin{equation}\label{eq:10}
    \begin{aligned}
        attr^{X_i}
        &\simeq A^{X_i}*G(A^{X_i})\\
        where\ G(A^{X_i})   &= \frac{\partial \theta(A)}{\partial A^{X_i}}
    \end{aligned}
\end{equation}

\begin{align}\label{eq:11}
    MIDAS
        &\simeq avg_t \{A^{X_1}*G_{norm}(A^{X_1})\notag\\
        & \quad \quad \quad - \sum_{X_0} (A^{X_0}*G_{norm}(A^{X_0})\notag\\
        \sum_n MIDAS &\simeq \mathbb{E}[A^{X_1}] - \sum_{X_0} \mathbb{E}[A^{X_0}]\\
    \text{where} & \quad G_{norm}(A) = \mathcal{N}(G(A))\notag\\
    & \quad \mathcal{N}(G(A)): G(A) \rightarrow (0,1)\notag
\end{align}

\subsection{LLM}
\subsubsection{Causal Objective: Implicit miATE maximization}
Before discussing LLM, we show that training a classifier implicitly addresses the $miATE$ maximization problem. Specifically, the objective (Eq. 6) over a hateful-confounder pair could be written as: 
\begin{equation} \label{eq:12}
        \mathop{\mathrm{argmin}}_\theta -\{f_{loss}(\theta_{X_1}) + \sum_{X_0} f_{loss}(1-\theta_{X_0})\}
\end{equation}
Here, we see that the first term maximizes the first term of Eq. 8 and the second term minimizes the second term of Eq. 8. In contrast, zero-shot LLM only addresses the first term of Eq. 13. In the next section, we show how we design the task for LLM to aim for the same goal.

\subsubsection{Causal Task Design}
To meta-optimize to the causal task, the hateful-confounder pair should be given \textit{to an identical, not separate, meta-optimizer}, or the optimizer cannot have any information about intersectional causality (second term of Eq. 12). Table 1 shows the causality-oriented design of a representative prompt for hateful meme detection.
\begin{table}[h!]
    \centering
    \begin{tabularx}{0.4\textwidth}{|X|}
        \hline
        User: Out of image-caption pairs \#0 to \#2, select the most likely hateful or sarcastic pair with a potential label (hateful or sarcastic). If all pairs are benign, please say so.\\ \#0: image: 'Skunk', caption: 'Love the way you smell today'\\
        \#1: image: 'Rose flower', caption: 'Love the way you smell today'\\
        \#2: image: 'Skunk', caption: 'Love the way skunks smell'\\
        System: \\
        \hline
    \end{tabularx}
    \caption{An illustrative prompt for the causal objective in the zero-shot scenario. In the few-shot context, answers are delivered succinctly (e.g., \textit{\#0 is hateful.}).}
    \label{table:1}
\end{table}

Note that this task design inherently counteracts sample imbalance since it simultaneously represents these hateful, original benign, and picked benign samples.

\subsubsection{Meta-Optimization for Causal Objective}
Meta-optimization for the causal objective poses challenges \cite{niuCounterfactualVQACauseEffect2021} like complicated instruction and varied available labels. The optimization process consists of:
\begin{enumerate}
    \item \textit{Task Type Classification (TTC)}. LLM recognizes the task as a binary classification.
    \item \textit{Label Identification (LI)}. LLM provides probable labels, e.g., \textit{hateful}, \textit{sarcastic} \cite{chauhanAllinOneDeepAttentive2020}, and \textit{benign}. Note that including the \textit{sarcastic} label addresses its nuanced overlap with hatefulness \cite{sundaramDistinguishingHateSpeech2022}, capitalizing on LLM's comprehension of complex social phenomena embedded in training corpora. We still regard this task as a binary classification of \textit{all-pair-benign} vs \textit{one-pair-hateful}, with a subtask of \textit{hateful-sarcastic} classification.
\end{enumerate}
See Table 2 for a set of examples.
\begin{table}[!ht]
    \centering
    \begin{tabularx}{\columnwidth}{|c|c|X|}
        \hline
        Subtask & Label & Response\\
        \hline
        TTC & Negative & Sorry, I couldn't understand your instructions.\\
        \hline
        TTC & Positive & \#1 could be sarcastic.\\
        \hline
        LI & Negative & \#1 could be sarcastic.\\
        \hline
        LI & Positive & \#0 could be hateful.\\
        \hline
    \end{tabularx}
    \caption{Synthesized responses and subtask labels for in-context learning. Refer to Table 1 for the corresponding instruction prompt. }\label{table.2}
\end{table}

The meta-optimization process is segmented into these subtasks, with the understanding that $LI$ follows a successful $TTC$, denoted as $TTC=1$.
The output of meta-optimized Transformer block $\theta$ could be formalized as:
\begin{equation}\label{eq:13}
    \begin{aligned}
        \theta^{SubTask} &= (A^{SubTask}+\Delta A^{SubTask})q\\
        \theta &= \left\{
            \begin{array}{ll}
                \theta^{TTC}+\theta^{LI} & (TTC = 1)\\
                \theta^{TTC} & otherwise
            \end{array}\right.\\
        where&\ SubTask \in \{TTC,LI\}
    \end{aligned}
\end{equation}

\section{Experimental Settings}
\subsection{Data Preparation}
\subsubsection{Hateful Memes Dataset} \label{dataset}
Our study utilizes the Hateful Memes Challenge dataset \cite{kielaHatefulMemesChallenge2020} and focuses primarily on the \textit{dev\_seen} subset. Unimodal hateful samples \cite{dasDetectingHateSpeech2020,lippeMultimodalFrameworkDetection2020a} are omitted from our study.

\subsubsection{Confounder Extraction} \label{conf}
From the dataset, 162 pairs of hateful $(T_1^{org}, I_1^{org})$ and benign samples $(T_1^{org}, I_0^{org})$ or $(T_0^{org}, I_1^{org})$ are identified. Since most of the pairs have either one of the text or image confounders, not both, three random inputs with the missing modality ($I_0^{picked}$ or $T_0^{picked}$) are concatenated with the other modality to accommodate the requirements of Eq. 8, resulting in a uniquely crafted subset $(T_1^{org}, I_0^{picked})$ and $(T_0^{picked}, I_1^{org})$. The structure of this subset is summarized in Table 3.

\begin{table}[!ht]
    \centering
    \begin{tabularx}{\columnwidth}{|X|c|}
        \hline
        % Sample Category & \makecell{Number of\\Samples}\\
        Sample Category & \# Samples\\
        \hline
        Hateful& 162\\
        \hline
        Image Benign& 78\\
        \hline
        Text Benign& 84\\
        \hline
        Picked Image Benign& 234\\
        \hline
        Picked Text Benign& 252\\
        \hline
    \end{tabularx}
    \caption{Samples utilized in our analysis. This table categorizes samples as Hateful, Image Benign $(T_1^{org}, I_0^{org})$, Text Benign $(T_0^{org}, I_1^{org})$, Picked Image Benign $(T_1^{org}, I_0^{picked})$, and Picked Text Benign $(T_0^{picked}, I_1^{org})$.}\label{table.3}
\end{table}

\subsection{Experiment I: Fine-Tuned Transformer} \label{Vilio}
\subsubsection{Analysis Type}
Assuming the predominant contribution of original inputs over the picked ones, in respect of the authors' effort of making the task challenging, we divided the analysis into that of $\{(T_1^{org}, I_1^{org}), (T_1^{org}, I_0^{org}), (T_0^{picked}, I_1^{org})\}$ (denoted as \textit{org. text}), and of $\{(T_1^{org}, I_1^{org}), (T_0^{org}, I_1^{org}), (T_1^{org}, I_0^{picked})\}$ (\textit{org. image}).
\subsubsection{Models}
We employ author implementation of the SOTA \cite{kielaHatefulMemesChallenge2021} \textit{Vilio} framework \cite{muennighoffVilioStateoftheartVisioLinguistic2020} for its superior capabilities and adaptable framework, focusing on its three main models: Oscar \cite{liOscarObjectSemanticsAligned2020}, UNITER \cite{chenUNITERUNiversalImageTExt2020}, and VisualBERT \cite{liVisualBERTSimplePerformant2020}, summarized in Table 4. Each model type has three submodels (training corpora or random seed variants), all included in our analysis but the results shown here are from selected one submodel (preliminary analysis shows all submodels exhibit similar trend).

\begin{table}[!ht]
    \centering
    \begin{tabularx}{\columnwidth}{|c|c|X|}
        \hline
        Type & Encoder & Pretraining Task\\
        \hline
        O & BERT (base) & 1) Object tag (or \textit{anchor}) detection 2) Text-image contrastive learning\\
        \hline
        U & BERT (base) & 1) Masked language modeling 2) Masked image modeling 3) Image-text matching 4) Word-region alignment via optimal transport\\
        \hline
        V & BERT (base) & 1) Masked language modeling 2) Image captioning \\
        \hline
    \end{tabularx}
    \caption{A categorization of Vilio's submodels leveraged in our research. The models are classified into three groups: Oscar (O), UNITER (U), and VisualBERT (V).}\label{table.4}
\end{table}

\subsection{Experiment II: LLM}\label{llm}
\subsubsection{Models}
HuggingFace \textit{Llama-2-13b-chat-hf} \cite{touvronLLaMAOpenEfficient2023} is our language model backbone, optimized for chat-style interactions. To convert the image into its textual description, we utilize the BLIP-2 \cite{liBLIP2BootstrappingLanguageImage2023} model with a \textit{FlanT5-XXL} \cite{chungScalingInstructionFinetunedLanguage2022} backbone.

\subsubsection{In-Context Learning}
We study Llama-2's behavior on image caption in the original dataset and image description generated by BLIP-2. For in-context learning, the number of samples is limited due to memory restriction. After the response is generated, one of the authors conducts manual labeling since the number of samples is limited (available at our GitHub repository). We gauge performance through accuracy.

\subsubsection{Meta-Gradient Evaluation} \label{meta}
During our evaluations, we mask redundant subtext (e.g., \textit{\#0: image:} and \textit{caption:}) in input prompts.

\subsection{Shared Settings}
\subsubsection{Probing}
We employ a probing \cite{alainUnderstandingIntermediateLayers2017} approach with LightGBM \cite{keLightGBMHighlyEfficient2017} to explore the impact of modality interaction and in-context learning on causal effect.  Responses are split into training (56\%), validation (14\%), and test (30\%) sets. We achieve hyperparameter tuning using Optuna \cite{akibaOptunaNextgenerationHyperparameter2019}. 
To assess the effects of interaction type $t$ (Experiment I, II) and model type (Experiment I), corresponding categorical variables and interaction terms with $MIDAS$ (Experiment I) or summed attention weights (Experiment II) are added to our analysis. We determine significance using a t-test ($p<0.05$).
\subsubsection{Text-Only Pretrained BERT}
For VisualBERT encoder replacement (Experiment I) and BLIP-2-fused-BERT (Experiment II), we use HuggingFace \textit{bert-base-uncased}. In Experiment II, the last four layers of BERT and a linear classifier are trained for 100 epochs with an Adam optimizer (learning rate 5e-5), evaluating its performance across different seeds.
\subsubsection{External Resources}
All code and experiments are accessible at\\
\url{https://github.com/HireTheHero/CausalIntersectionalityDualGradient}.\\Experiments are conducted on a single NVIDIA A100 GPU, either through Google Colaboratory Pro+ or locally.

\section{Results \& Discussion}
\subsection{Experiment I: Fine-Tuned Transformer}
\subsubsection{miATE}
First, we assessed each model's performance with $miATE$ (Fig. 3). VisualBERT exhibited the highest disparity, highlighting its bias for text-based tasks (Table 4).

\begin{figure}[!ht]
    \begin{center}
    %\fbox{\parbox{6cm}{
    %This is a figure with a caption.}}
    % old picture \includegraphics[scale=0.5]{lrec2020W-image1.eps} 
    \includegraphics[scale=0.5]{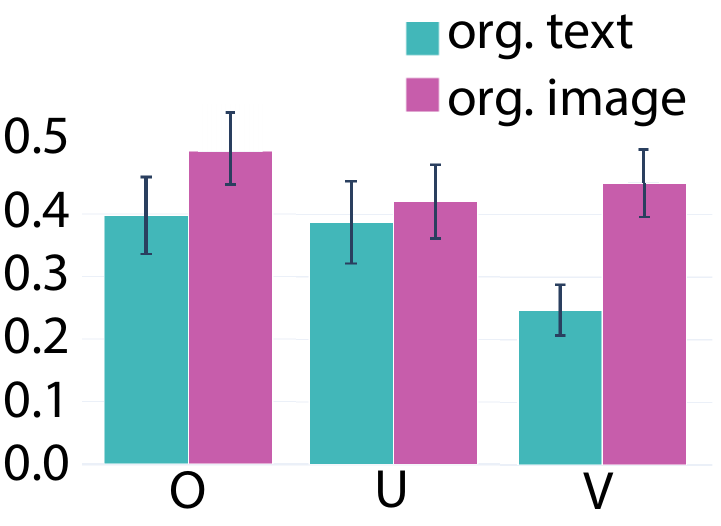} 

    \caption{Multimodal Intersectional Average Treatment Effect ($miATE$) across Oscar (O: left), UNITER (U; middle), and VisualBERT (V; right) models, contrasting the samples with original image confounders (\textit{org. image}, cyan) and those with original text confounders (\textit{org. text}, magenta).}
    \label{fig.3}
    \end{center}
\end{figure}

\subsubsection{MIDAS Global Analysis}
Next, we assessed the model's inner workings (Fig. 4 and 5). $MIDAS$ of Oscar and UNITER (Fig. 4, first and second row) showed predictable trends of attending to one modality while the other is the same. In contrast, VisualBERT's behavior of attending to text-related interactions (third row) mirrored its pretraining tendencies biased towards text (Table 4). Furthermore, replacing VisualBERT's encoder with the one pretrained only with text enhanced the bias (fourth row), which supports the presence of pretraining bias represented in $MIDAS$. We observed no significant model differentiation with the original $attr$ (Fig. 4, left column of each graph), suggesting a simple yet important contribution of modality-wise split.

\begin{figure}[!ht]
    \begin{center}
    %\fbox{\parbox{6cm}{
    %This is a figure with a caption.}}
    % old picture \includegraphics[scale=0.5]{lrec2020W-image1.eps} 
    \includegraphics[scale=0.5]{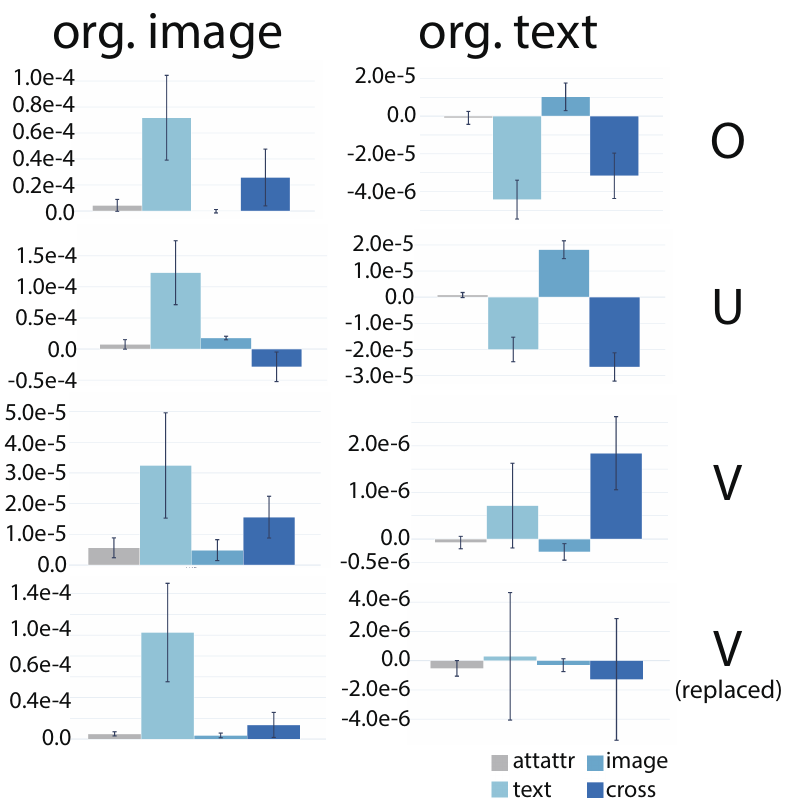} 

    \caption{$MIDAS$ for \textit{org. image} (left) and \textit{org. text} (right) samples featuring Oscar (top), UNITER (second row), VisualBERT (third row), and VisualBERT with text-only-pretrained encoder (bottom). From left to right, each graph displays $attr$ with no modality division, $MIDAS_{within\_text}$, $MIDAS_{within\_image}$, and $MIDAS_{cross\_modal}$.}
    \label{fig.4}
    \end{center}
\end{figure}

\subsubsection{MIDAS Local Analysis}
To see if we can interpret the single hateful-benign pair, we extracted local explanation \cite{chaiExplainableMultiModalHierarchical2021,heeExplainingMultimodalHateful2022}. A representative pair (Fig. 5) illustrates that UNITER captures the contrast between a woman and cargo in image confounder analysis (first and third row), and the model similarly attended to the words \textit{dishwasher} and \textit{driving} for text analysis (first and second row).

\begin{figure}[!ht]
    \begin{center}
    %\fbox{\parbox{6cm}{
    %This is a figure with a caption.}}
    % old picture \includegraphics[scale=0.5]{lrec2020W-image1.eps} 
    \includegraphics[scale=0.5]{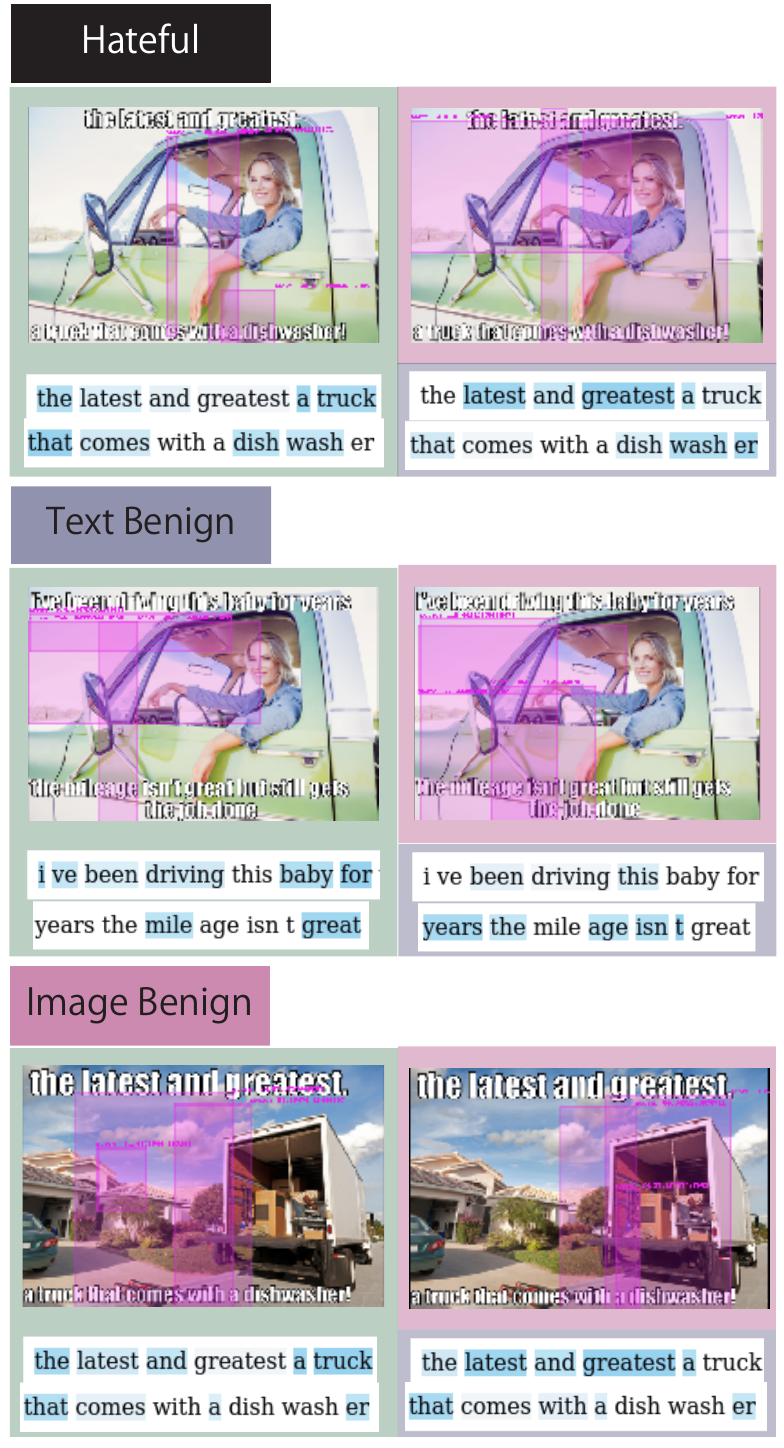} 

    \caption{Conceptual portrayal of hateful, text benign, and image benign samples derived from UNITER. $MIDAS$ reflects heightened $attr_{cross\_modal}$ (green), $attr_{within\_image}$ (red), or $attr_{within\_text}$ (blue) values. Both image and text inputs spotlight top-scored ROIs and tokens. The text is abbreviated for clarity.}
    \label{fig.5}
    \end{center}
\end{figure}

\subsubsection{Empirical Relation between MIDAS \& miATE}
To probe the relationship between $MIDAS$ and $miATE$, we first modeled the entire $(MIDAS,miATE)$ pairs of all the models by a single probe, resulting in a moderate AUC of $75.6\pm4.20$. Next, to see the effectiveness of probing for each model, we applied one probe for one model, resulting in the highest AUC for V (AUC $94.1\pm3.01$), while low-to-moderate for O ($60.8\pm5.18$) and U ($74.3\pm2.39$). In addition, to see if the probes reflect the findings on $MIDAS$, we analyzed the feature importance (Table 5). Consistent with previous findings, VisualBERT and within-text appear with the highest frequency among model type and interaction type, respectively. In summary, our findings showed a moderate correlation between $MIDAS$ and $miATE$ for BERT-based models, with a particularly robust link for VisualBERT, echoing its distinct model nature.

\begin{table}
    \centering
    \begin{tabular}{|p{2.5cm}|p{1cm}|p{1cm}|p{1.3cm}|}
        \toprule
        Interaction Type    & O& U&V\\
        \midrule
        within-text        & 35$\pm$25& 30$\pm$26&253$\pm$66\\
        cross-modal      & 19$\pm$14& 35$\pm$22&223$\pm$75\\
        within-image    & 30$\pm$23& 54$\pm$32&169$\pm$74\\
        \bottomrule
    \end{tabular}
    \caption{LightGBM probe's feature importance between $MIDAS$ and $miATE$. Values represent frequency counts.}\label{table.6}
\end{table}

\subsection{Experiment II: LLM}
\subsubsection{Effectiveness of BLIP-2 information retrieval}\label{blip2}
To assess the merit of BLIP-2 information retrieval, we utilized its image description and the original captions to fine-tune BERT pretrained only with text. The resultant enhanced performance (Accuracy $66.9\pm0.84$, AUC: $71.2\pm1.55$) to unimodal BERT benchmarks \cite{kielaHatefulMemesChallenge2020} underlines BLIP-2's effective image information extraction capabilities.

\subsubsection{In-Context LLM Performance}\label{icl_result}
Our evaluation with Llama-2 shows that all-sample accuracy improved after one sample (Table 6, left). Interestingly, after just one in-context example, the model achieved exactly the same performance for all samples and $TTC=1$ samples, meaning impeccable TTC Recall. These results suggest the critical role of in-context examples in task comprehension when the task is challenging in zero-shot settings. Marking the sarcastic label as positive led to better performance at the zero-shot setting but dropped after one example (Table 6, middle and right), implying uncertainty in the decision-making for this label.

\begin{table}
    \centering
    \begin{tabular}{|c|c|c|c|c|}
        \hline
        \multicolumn{1}{|c|}{\# Few-Shot} & \multicolumn{3}{c|}{Accuracy}\\
        \cline{2-4}
        \multicolumn{1}{|c|}{Samples} & All(S-) & TTC(S-) & TTC(S+)\\
        \hline
        0   &46.2   &61.6   &62.6\\
        1   &62.9   &62.9   &60.6\\
        2   &62.5   &62.5   &61.0\\
        3   &59.2   &59.2   &57.6\\
        4   &64.3   &64.3   &71.4\\
        \hline
    \end{tabular}
    \caption{Zero-shot (first row) and few-shot (second to fifth row) Llama-2 performance. \textit{All} signifies cumulative sample results, while \textit{TTC} relates to correct TTC samples. Parentheticals (\textit{S+} or \textit{S-}) denote the inclusion of the \textit{sarcastic} label in either positive or negative samples.} \label{table.7}
\end{table}

\subsubsection{Meta-Optimization Evaluation}\label{meta_result}
To gauge the influence of meta-optimization, we applied a probe model to examine the relationship between summed attention weights $(A, \Delta A)$ and TTC label, revealing a moderate AUC of $82.3\pm6.16$. Furthermore, a detailed extraction of feature importance (Table 7) from the probe model allowed us to determine how $A$ and $\Delta A$ impact the probe model (left and right). Our findings suggest that while 
$A$ carries substantial weight in decision-making, its meta-optimized counterpart $\Delta A$ also plays a vital role. Regarding the effects of modality interaction (top and bottom), captured by interaction-type-divided weights $A+\Delta A$, each interaction type did contribute to TTC. When examining the differential impacts of each type, however, no significant disparities in their contributions were identified. Addressing the challenge of discerning between them will be a part of the future work.
\begin{table}
    \centering
    \begin{tabular}{|p{2.5cm}|p{1cm}|p{1cm}|}
        \toprule
        Interaction Type    & $A$& $\Delta A$\\
        \midrule
        within-text        & 65$\pm$41& 24$\pm$17\\
        cross-modal      & 58$\pm$33& 31$\pm$21\\
        within-image    & 43$\pm$25& 14$\pm$8\\
        \bottomrule
    \end{tabular}
    \caption{LightGBM probe's feature importance in Experiment II. Features of zero-shot weights ($A$) or their few-shot updates ($\Delta A$) are divided by interaction types.}\label{table.9}
\end{table}

\subsection{Discussion}\label{discussion}
The primary goal of this paper is to assess models based on the data generation process and its underlying concepts - \textit{hatefulness} in the case of hateful memes. While this deviates from standard ML evaluations focusing on performance metrics like accuracy, it is scientifically valid and relevant to ML problems, like Rubin started his line of causal inference works to analyze the impact of nulled variables \cite{rubinObjectiveCausalInference2008}.  
In our study, we demonstrate that the generation of hateful memes embodies multimodal intersectionality, and the SOTA Transformer models effectively capture this nature of the data but are biased by pretraining datasets. In the future, we hope to apply our method to other multimodal problems like the missing modality problem \cite{maSMILMultimodalLearning2021,wangMultiModalLearningMissing2023}, an inherently close one to nulled variable evaluation.\\
Our study carves a niche by reconceptualizing hateful meme detection through the lens of modality interaction and causal effect. Compared to the seminal work on causal intersectionality \cite{brightCausallyInterpretingIntersectionality2016}, beyond mere technical insights, we proffer a paradigm shift in causal intersectionality.\\
Our method's key advantage is its unique capability for modality-wise causal analysis, a novel contribution in this field. Despite its simplicity, the causal effect of the modalities is neither investigated nor formally defined in the existing literature. \\
Empirically, Experiment I unveils model biases overlooked by traditional methods like attention attribution scores \cite{haoSelfAttentionAttributionInterpreting2021,heeExplainingMultimodalHateful2022} without consideration for modality.\\
For Experiment II, our exploration of few-shot LLM performance has provided an understanding of how LLM adapts to different levels of input information, shedding light on their capabilities and limitations in various scenarios. Applying meta-gradients has allowed us to assess the attribution of attention weights, adding granularity to the interpretability landscape. Evaluating the effectiveness of the causal task over the causal evaluation of LLMs is challenging since it is a new concept. Nonetheless, this could be a valuable benchmark for future model evaluations. Despite relying on a specific instruction prompt for the causal task, we could adapt the design for broader applications. For example, with a simple modification to the prompt, we could test LLMs with multi-class meme classification \cite{davidsonAutomatedHateSpeech2017}.

\section{Conclusion}\label{conclusion}
We posit that hateful meme detection transcends mere classification, gravitating towards intersectional causal effect analysis. Our evaluations spanned various Transformer architectures in unique settings. To ensure our approach's universality, extending our evaluations to other hateful memes datasets \cite{gomezExploringHateSpeech2020,dasHateMMMultiModalDataset2023} will be pivotal. In the quest for broader insights, exploring diverse challenges, such as the intersectionality in multimodal medical analyses \cite{azilinonCombiningSodiumMRI2023}, will be part of our future work. For scalability, utilizing more of the power of LLMs will be promising for confounder extraction and response evaluation.

\newpage

\section{Ethical Considerations}
In this research, we aim to develop innovative analytical methods for identifying and mitigating the proliferation of hateful memes, a pressing concern given the complex interplay of text and imagery in propagating hate speech online. The nature of this endeavor necessitates rigorous ethical scrutiny, especially concerning the selection, utilization, and presentation of these memes within our scholarly work and its broader dissemination. Herein, we delineate the principal ethical considerations guiding our study.\\
The hateful memes we examine originate from a prior investigation \cite{kielaHatefulMemesChallenge2020}. We direct readers to the original study for insights into the ethical measures employed during the dataset's compilation. Our engagement with this dataset is underpinned by a firm commitment not to propagate or validate the adverse messages it encompasses.\\
Our sample selection approach is predicated on a causality framework detailed in the \textit{Confounder Extraction} section, ensuring a comprehensive examination of hate speech manifestations across diverse community targets. This methodology underscores our commitment to a nuanced analysis that refrains from generalizations or biases.\\
A pivotal aspect of our ethical strategy is to reconcile the imperative of methodological transparency with the necessity to limit harm. Consequently, we exhibit restraint in our presentation of hateful memes. Specifically, Figure 5 is the sole instance within our publication where an actual hateful meme is depicted. We have exercised meticulous care to ensure that neither the accompanying text description nor the figure caption disseminates any form of hate speech.\\
This strategy is emblematic of our broader ethical stance, emphasizing the conscientious handling of sensitive content. Our research is animated by a profound dedication to combating hate speech in all its forms, reflecting an unwavering commitment to ethical research practices that respect the dignity of all individuals and communities. Through this work, we aspire not only to advance the field of hate speech detection but also to contribute meaningfully to creating more inclusive and respectful digital spaces.

\section{Limitations}
This study's primary limitation concerns the unverified generalizability of its findings. Hateful memes represent an evolving area of concern that necessitates extensive, openly accessible datasets for comprehensive analysis and validation. Our research endeavors to tackle this challenge, yet the broader scope for future exploration is highlighted by the potential applications of our findings, as detailed in the \textit{Discussion} and the \textit{Conclusion} sections.\\
A further constraint is the linguistic homogeneity of the dataset employed, with the Hateful Memes Challenge dataset comprising exclusively English-language textual content. This presents a critical limitation in the context of the global escalation of extremism, where hate speech proliferates across linguistic boundaries. The detection of multilingual hate speech thus emerges as a crucial area for future research, necessitating methodologies capable of navigating language-specific nuances and cultural contexts.\\
Additionally, the field of hate speech detection faces resource limitations, notably in the size and diversity of available datasets. Hateful speech datasets are generally small, restricting the depth and breadth of training data for machine learning models. We believe future studies could utilize LLMs as dataset curators.\\
In summary, while this study contributes valuable insights into detecting and mitigating hateful memes, it also underscores the need for further research. Addressing the limitations related to dataset generalizability, linguistic diversity, and the scarcity of training data are pivotal steps toward developing more effective and universally applicable solutions for combating online hate speech. Exploring innovative methods, such as LLM-based dataset curation, represents a promising direction for overcoming these challenges.\\
From a theoretical standpoint, the groundwork of our in-context learning analysis relies upon the principles of simplified linear attention \cite{irieDualFormNeural2022,daiWhyCanGPT2023a}. However, this foundation's direct applicability to conventional Transformer models invites scrutiny. Consequently, a more nuanced interpretation \cite{renIncontextLearningTransformer2023} may be imperative for advancing our understanding in future investigations.

\section{Acknowledgments}
CyberAgent Inc.\footnote{\url{https://www.cyberagent.co.jp/}} supported this work, enabling our conference participation. We are profoundly grateful to the anonymous reviewers, whose constructive feedback significantly enriched our discussions and contributed to refining our research. Yosuke Miyanishi thanks J.X., H.W., and T.Y. for their insightful comments that significantly improved our study.

% \begin{figure}[!ht]
% \begin{center}
% %\fbox{\parbox{6cm}{
% %This is a figure with a caption.}}
% % old picture \includegraphics[scale=0.5]{lrec2020W-image1.eps} 
% \includegraphics[scale=0.5]{turin2024-banner.jpg} 

% \caption{The caption of the figure.}
% \label{fig.1}
% \end{center}
% \end{figure}

% \begin{table}[!h]
% \begin{center}
% \begin{tabularx}{\columnwidth}{|l|X|}

%       \hline
%       Level&Tools\\
%       \hline
%       Morphology & Pitrat Analyser\\
%       \hline
%       Syntax & LFG Analyser (C-Structure)\\
%       \hline
%      Semantics & LFG F-Structures + Sowa's\\
%      & Conceptual Graphs\\
%       \hline

% \end{tabularx}
% \caption{The caption of the table}
%  \end{center}
% \end{table}

%\subsection{Big tables}
%
%An example of a big table which extends beyond the column and will
%float in the next page.
%
% \begin{table*}[ht]
% \begin{center}
% \begin{tabular}{|l|l|}
%
%       \hline
%       Level&Tools\\
%       \hline\hline
%       Morphology & Pitrat Analyser\\
%       Syntax & LFG Analyser (C-Structure)\\
%       Semantics & LFG F-Structures + Sowa's Conceptual Graphs  \\
%       \hline
%
% \end{tabular}
% \caption{The caption of the big table}
% \end{center}
% \end{table*}
%

% \section{References}\label{reference}
%\label{main:ref}

\clearpage
\bibliography{aaai24}

\appendix
\section{Appendix}
\subsection{Further Exploration for Local Explainability}
Sample analysis for Oscar (Fig. 6) shows a similar trend to UNITER (Fig. 5). Interestingly, VisualBERT does not attend to the key components (woman or cargo) in the image, supporting its bias towards textual information.
\begin{figure}[!ht]
    \begin{center}
    %\fbox{\parbox{6cm}{
    %This is a figure with a caption.}}
    % old picture \includegraphics[scale=0.5]{lrec2020W-image1.eps} 
    \includegraphics[scale=0.7]{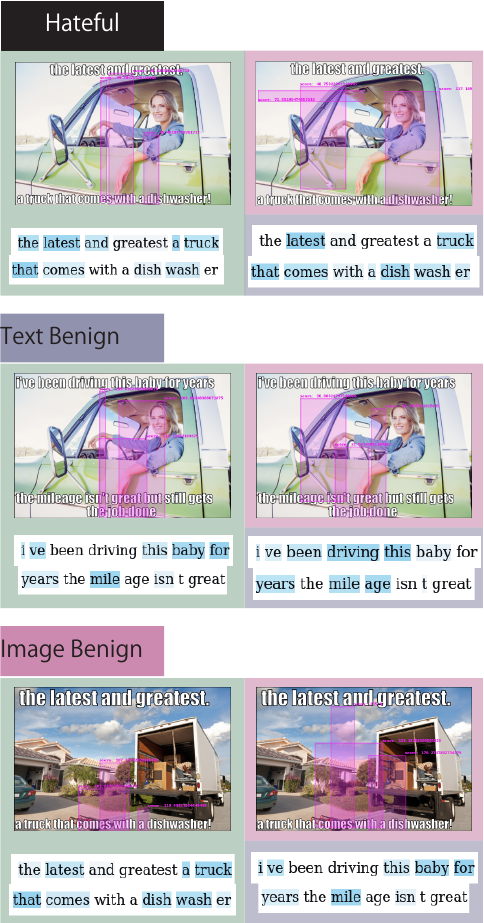} 

    \caption{Sampled derived from Oscar.}
    \label{fig.6}
    \end{center}
\end{figure}
\begin{figure}[!ht]
    \begin{center}
    %\fbox{\parbox{6cm}{
    %This is a figure with a caption.}}
    % old picture \includegraphics[scale=0.5]{lrec2020W-image1.eps} 
    \includegraphics[scale=0.7]{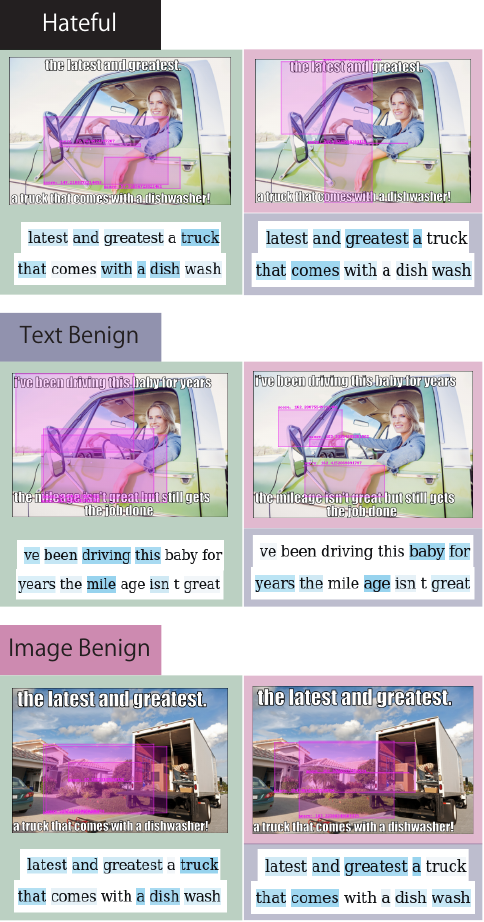} 

    \caption{Samples derived from VisualBERT.}
    \label{fig.7}
    \end{center}
\end{figure}
\newpage
\subsection{Breakdown of Attention Attribution Score}
The attention attribution score $attr$ (Eq. 3) is the product of the attention weight matrix and the integral of the gradient. To see the separate impact, we replaced the $attr$ term of the $MIDAS$ equation (Eq. 9) with the attention $MIDAS_{att}$ or the gradient $MIDAS_{grad}$ for comparison. In general, $MIDAS_{att}$ (Fig. 8-10) shows a more similar trend to the original $MIDAS$ than $MIDAS_{grad}$ (Fig. 11-13). This result implies that the attention weights decide the model's strategy, while the gradient adjusts the impact of the individual component.
\begin{figure}[!ht]
    \begin{center}
    %\fbox{\parbox{6cm}{
    %This is a figure with a caption.}}
    % old picture \includegraphics[scale=0.5]{lrec2020W-image1.eps} 
    \includegraphics[scale=0.7]{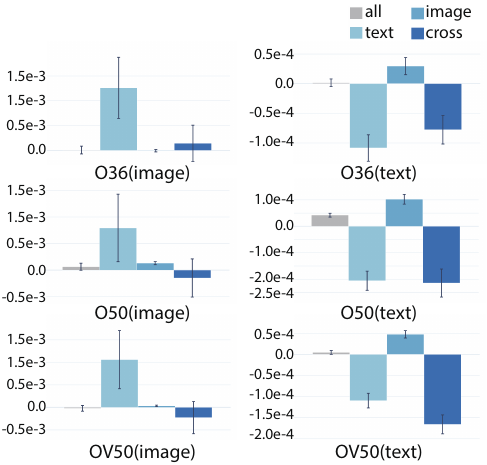} 

    \caption{Oscar $MIDAS_{att}$.}
    \label{fig.8}
    \end{center}
\end{figure}
\begin{figure}[!ht]
    \begin{center}
    %\fbox{\parbox{6cm}{
    %This is a figure with a caption.}}
    % old picture \includegraphics[scale=0.5]{lrec2020W-image1.eps} 
    \includegraphics[scale=0.7]{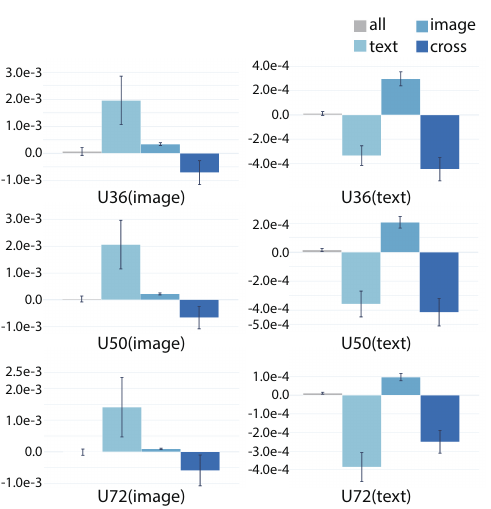} 

    \caption{UNITER $MIDAS_{att}$.}
    \label{fig.9}
    \end{center}
\end{figure}
\begin{figure}[!ht]
    \begin{center}
    %\fbox{\parbox{6cm}{
    %This is a figure with a caption.}}
    % old picture \includegraphics[scale=0.5]{lrec2020W-image1.eps} 
    \includegraphics[scale=0.7]{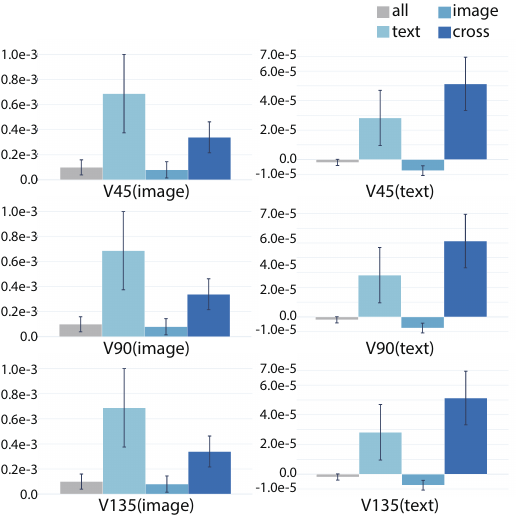} 

    \caption{VisusalBERT $MIDAS_{att}$.}
    \label{fig.10}
    \end{center}
\end{figure}
\begin{figure}[!ht]
    \begin{center}
    %\fbox{\parbox{6cm}{
    %This is a figure with a caption.}}
    % old picture \includegraphics[scale=0.5]{lrec2020W-image1.eps} 
    \includegraphics[scale=0.7]{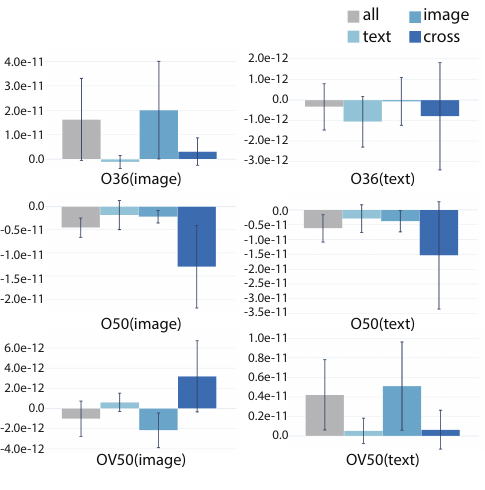} 

    \caption{Oscar $MIDAS_{grad}$.}
    \label{fig.11}
    \end{center}
\end{figure}
\begin{figure}[!ht]
    \begin{center}
    %\fbox{\parbox{6cm}{
    %This is a figure with a caption.}}
    % old picture \includegraphics[scale=0.5]{lrec2020W-image1.eps} 
    \includegraphics[scale=0.7]{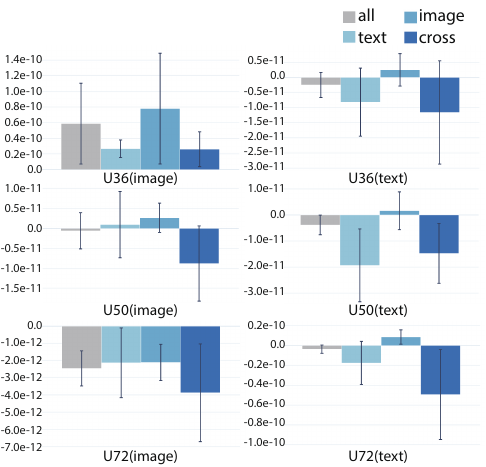} 

    \caption{UNITER $MIDAS_{grad}$.}
    \label{fig.12}
    \end{center}
\end{figure}
\begin{figure}[!ht]
    \begin{center}
    %\fbox{\parbox{6cm}{
    %This is a figure with a caption.}}
    % old picture \includegraphics[scale=0.5]{lrec2020W-image1.eps} 
    \includegraphics[scale=0.7]{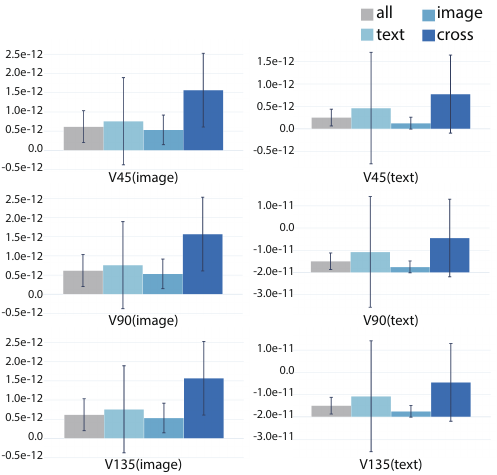} 

    \caption{VisusalBERT $MIDAS_{grad}$.}
    \label{fig.13}
    \end{center}
\end{figure}
\end{document}